\documentclass{article} % For LaTeX2e
\usepackage[final]{colm2026_conference}
\usepackage{microtype}
\usepackage{hyperref}
\usepackage{url}
\usepackage{booktabs}
\usepackage{amsmath}
\usepackage{multirow}
\usepackage{multicol}
\usepackage[table]{xcolor}
\usepackage{graphicx}
\usepackage{subcaption}
\usepackage{caption}
\usepackage{svg}
\usepackage{amsmath,amssymb,amsthm}

\usepackage{adjustbox}
\usepackage{wrapfig}
\definecolor{rlcmshade}{RGB}{242,247,253}

\newcommand{\best}[1]{\textbf{#1}}
\newcommand{\second}[1]{\underline{#1}}

\newcommand{\promptbox}[1]{
\begin{center}
\fcolorbox{darkblue}{rlcmshade}{
\begin{minipage}{0.94\linewidth}
\small
#1
\end{minipage}}
\end{center}
}

\usepackage{lineno}
\definecolor{darkblue}{rgb}{0, 0, 0.5}
\hypersetup{colorlinks=true, citecolor=darkblue, linkcolor=darkblue, urlcolor=darkblue}

\title{Process Supervision of Confidence Margin for Calibrated LLM Reasoning}

\author{
    \textbf{Liaoyaqi Wang}{\hspace{.1em}}
    \quad
    \textbf{Chunsheng Zuo}{\hspace{.1em}}
    \quad
    \textbf{William Jurayj}{\hspace{.1em}}
    \quad
    \textbf{Benjamin Van Durme}{\hspace{.1em}}
    \quad
     \textbf{Anqi Liu}{\hspace{.1em}}
    \vspace{.5em}\\
    Johns Hopkins University
    \vspace{.5em}\\
    \texttt{\{lwang240,czuo3,wjurayj1,bvandur1,aliu.cs\}@jh.edu}
}

\newcommand{\method}[1]{{\color{blue} RLCM}}

\begin{document}

\ifcolmsubmission
\linenumbers
\fi

\maketitle

\begin{abstract}
Scaling test-time computation with reinforcement learning (RL) has emerged as a reliable path to improve large language models (LLM) reasoning ability. Yet, outcome-based reward often incentivizes models to be overconfident, leading to hallucinations, unreliable confidence-based control, and unnecessary compute allocation. 
We introduce Reinforcement Learning with Confidence Margin (\textbf{RLCM}), a calibration-aware RL framework that jointly optimizes correctness and confidence reliability via a margin-enhanced process reward over intermediate-budget completions. 
Rather than aligning confidence to correctness likelihoods, RLCM encourages the model to widen the confidence margin between correct and incorrect steps within a single reasoning trajectory.
Across mathematical, code, logic and science benchmarks,
our method substantially improves calibration while maintaining or improving accuracy. 
We further show that, with calibrated confidence signals, the resulting models enable more efficient conformal risk control and effective confidence-weighted aggregation\footnote{\url{https://github.com/Melodramass/RLCM}}\footnote{\url{https://huggingface.co/collections/Melodramas/rlcm}}. 
\end{abstract}

\section{Introduction}

Reasoning large language models (LLMs) can solve substantially harder problems by scaling test-time computation~\citep{snell2025scaling,muennighoff-etal-2025-s1}, achieving impressive gains across domains such as mathematics, coding and scientific discovery~\citep{shao2024deepseekmathpushinglimitsmathematical,openai2024openaio1card,deepseekai2025deepseekr1incentivizingreasoningcapability}. \begin{wrapfigure}[18]{r}{0.41\textwidth}
    \centering
    \includegraphics[width=\linewidth]{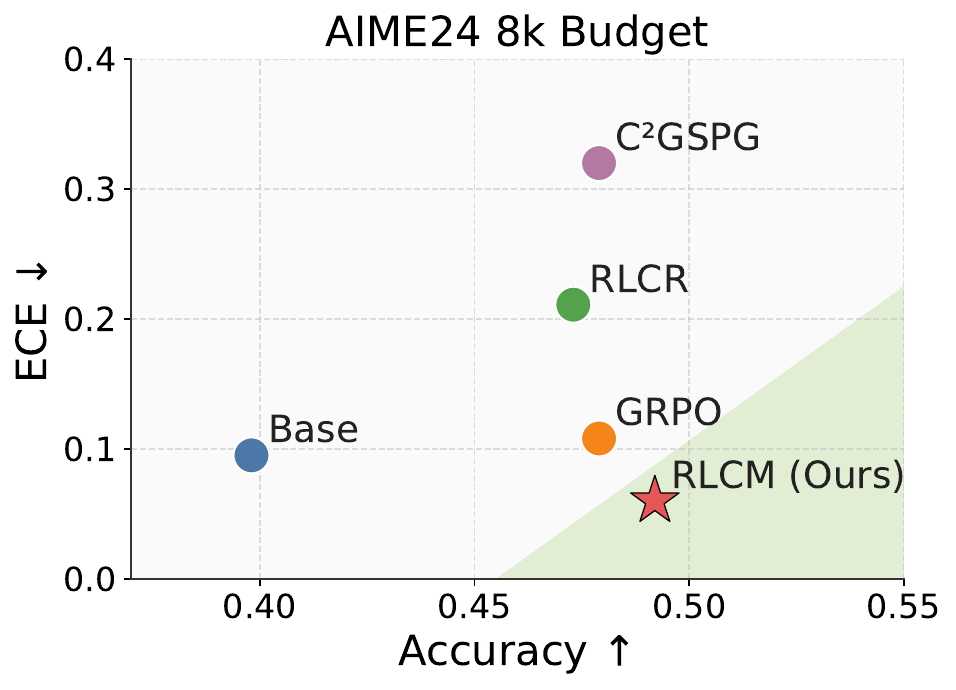}
    \caption{
 \small{   Accuracy v.s. ECE at the 8k token budget on AIME24. RLCM preserves high accuracy while substantially improving calibration, achieving the strongest Pareto trade-off among the compared methods.}
    }
    \label{fig:teaser_fig}
\end{wrapfigure} However, strong reasoning performance does not guarantee reliable confidence expression~\citep{10.1145/3703155,yona-etal-2024-large,jurayj-etal-2025-final}.
This limitation is partly rooted in outcome-driven training objectives, which optimize final answer correctness but do not explicitly encourage the model's confidence to reflect its true probability of being correct~\citep{kalai2025why,mei2026reasoning}.

Moreover, because supervision is concentrated on the final outcome, these objectives provide only weak guidance for how confidence should evolve throughout the reasoning process~\citep{lightman2024lets,luo2025improve}.
As a result, reasoning models may be confidently wrong, and their expressed uncertainty may not faithfully reflect the quality of the reasoning process and final answers~\citep{lacombe2025think}. \autoref{fig:teaser_fig} shows that existing methods struggle to jointly optimize accuracy and calibration, measured by accuracy and expected calibration error (ECE).

Misplaced confidence hinders deployment in high-stakes contexts.
Uncertainty is a widely used upstream signal for decision making: it can govern whether the model should stop early, revise its answer, invoke external tools, or defer to human experts~\citep{wu-etal-2025-thought,jiang2025conformal,ni-etal-2024-llms,10.1609/aaai.v39i27.35063}.
Unreliable confidence therefore limits the safe deployment of reasoning models, especially in high-stakes domains and makes it difficult to steer model behavior through uncertainty~\citep{10.5555/3692070.3692180,10.1145/3711896.3736569}. 
Existing works training language models to express uncertainty, whether based on answer consistency \citep{huang_efficient_2025}, verbalized confidence \citep{leng2025taming}, or post-hoc probing \citep{zhang2025reasoning}, largely focus on final-answer uncertainty and yield limited control over how confidence evolves across intermediate reasoning steps, leaving process-level calibration largely underexplored.

\begin{figure}
\small
    \centering
    \includegraphics[width=0.9\linewidth]{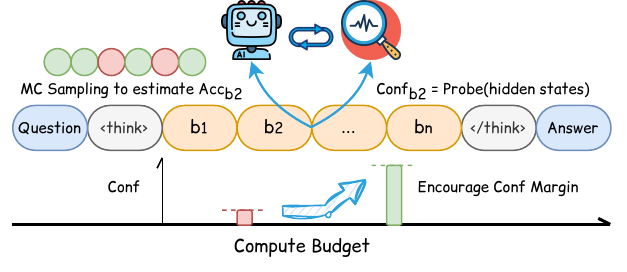}
    \caption{
    Overview of \textbf{RLCM}. We select intermediate compute budgets \(b_1,\dots,b_n\) along a reasoning trajectory. At each budget, Monte Carlo sampling estimates intermediate correctness, while a lightweight probe predicts confidence from the hidden state. Instead of directly aligning confidence with correctness, RLCM encourages a larger confidence margin between correct and incorrect intermediate states within the same trajectory.}
    \label{fig:figure1}
\end{figure}

To address this, we propose \textbf{RLCM} (Reinforcement Learning with Confidence Margin), a calibration-aware RL framework that supervises confidence throughout the reasoning trajectory with a margin-based process reward. As is shown in the \autoref{fig:figure1},
we incentivize a model to increase the confidence gap~\citep{chhikara2025mind} between correct and incorrect intermediate reasoning steps within the same trajectory, providing more useful scores for downstream calibration.
To make such supervision available during training, we attach a lightweight confidence probe to intermediate reasoning states and use it to estimate how likely a truncated reasoning prefix is to generate a correct final answer. This relative margin-enhanced objective provides a more robust calibration signal than direct score matching while remaining compatible with final-answer optimization. 
By imposing this margin reward at intermediate reasoning steps during the on-policy training stage, we facilitate flexible and cheap post-hoc calibration to address diverse efficiency and safety objectives. 

Across mathematical reasoning benchmarks of varying difficulty, and on out-of-domain tasks such as coding, scientific, and logic question answering, RLCM improves calibration while preserving or even improving reasoning accuracy, as we indicated in the \autoref{fig:teaser_fig}. These improvements also translate to downstream decision making: calibrated confidence enables more effective conformal risk control, which enforces target error guarantees, and stronger confidence-weighted aggregation results.

\section{Background and Related Work}

\subsection{Language Model Reinforcement Learning}
\textbf{Reinforcement Learning with Verifiable Rewards.} Recent progress in reasoning language models has been driven by reinforcement learning with verifiable rewards (RLVR) \citep{lambert_tulu_2025}, exemplified by Group Relative Policy Optimization (GRPO) \citep{shao2024deepseekmathpushinglimitsmathematical, deepseekai2025deepseekr1incentivizingreasoningcapability}. Given a prompt \(x\), the policy \(\pi_\theta\) samples a group of completions $\{y_1,y_2,...y_i\}$, each receiving a scalar reward \(R_i\) by verifying its predicted answer with the ground-truth answer. GRPO optimizes: 
{\small
\[
\begin{aligned}
\mathcal{J}_{\mathrm{GRPO}}(\theta)=
\mathbb{E}_{x,y}
\Bigg[
\frac{1}{G}\sum_{i=1}^{G}\frac{1}{|y_i|}\sum_{t=1}^{|y_i|}
\min\!\Big(
r_{i,t}(\theta)\hat{A}_i,\;
\mathrm{clip}\!\big(r_{i,t}(\theta),1-\epsilon,1+\epsilon\big)\hat{A}_i
\Big)
-\beta\,D_{\mathrm{KL}}\!\left(\pi_\theta \,\|\, \pi_{\mathrm{ref}}\right)
\Bigg],
\end{aligned}
\]
}
where
{\small
\[
r_{i,t}(\theta)
=
\frac{\pi_\theta(y_{i,t}\mid x,y_{i,<t})}
{\pi_{\theta_{\mathrm{old}}}(y_{i,t}\mid x,y_{i,<t})},
\qquad
\hat{A}_i
=
\frac{R_i-\mathrm{mean}(\{R_j\}_{j=1}^G)}
{\mathrm{std}(\{R_j\}_{j=1}^G)}.
\]
}

By using a group-relative advantage as a surrogate for the value function, GRPO training is more efficient, obviating the need for a separate value model like in PPO \citep{schulman2017proximal}. 
However, optimizing exclusively for verifiable correctness often triggers \textbf{calibration degeneration}~\citep{luo2025your}, as binary rewards incentivize models to prioritize correctness over honesty~\citep{kalai2025why,bereket2025calibration}. This necessitates training strategies that harmonize outcome rewards with internal uncertainty, motivating our design of an auxiliary calibration reward.
We also remove the KL regularizer following \citep{yu2025dapo, zhang2026designkl}.

\textbf{Process Rewards.} RLVR suffers from sparse supervision since rewards are usually assigned only at the end of a rollout. To improve credit assignment, recent work introduces process rewards that provide denser intermediate feedback~\citep{zheng2025surveyprocessrewardmodels}. Such rewards can be derived from human annotations or LLM judgments for mathematical reasoning \citep{yu_ovm_2024, wang_math-shepherd_2024, zhang_lessons_2025}, from LLM-based evaluation of tool utility in agentic tasks \citep{goldie_synthetic_2025}, or from Monte Carlo completions sampled from intermediate states \citep{li2025enhancingreasoningprocesssupervision, qi2025optimizing, qu2025optimizing}. Another line of work uses stronger models to score or distill token-level reasoning traces, including on-policy distillation \citep{agarwal_-policy_2023}, 
and recent self-distillation variants \citep{zhao_self-distilled_2026, hubotter_reinforcement_2026}. These approaches, however, often require costly external model calls. In contrast, RLCM relies on a lightweight probe to deliver efficient process-level supervision.

\subsection{Uncertainty and Calibration in LLMs}
\textbf{Uncertainty Estimation.} Transformer-based language models are often miscalibrated and tend to be overconfident in their predictions, which undermines the reliability of their reported confidence \citep{guo2017calibration,desai2020calibration,jiang2022calibrating}. To improve uncertainty estimation quality in LLMs, prior work has explored a broad range of approaches, including logit- or probability-based \citep{pmlr-v139-zhao21c,gupta2024language}, entropy- and consistency-based signals derived from sampled generations \citep{farquhar2024detecting,fu2025deepthinkconfidence,kang2025scalable}, verbalized confidence elicitation \citep{tian-etal-2023-just,xiong2024can,wang2026calibrating}, probing internal representations \citep{ch-wang-etal-2024-androids,kossen2025semantic}, and directly training models to predict calibrated probabilities \citep{chen-etal-2020-uncertain,wang2025always}. Probe-based methods are especially promising for fine-grained uncertainty estimation, but their gains are often strongest in-domain, and robustness under distribution shift remains a central challenge~\citep{bakman-etal-2025-reconsidering}.

\textbf{Calibration and Control.} Existing works approach post-hoc LLM calibration from several directions. Among post-hoc approaches, conformal prediction is particularly appealing because it provides distribution-free guarantees for prediction sets, intervals, or abstention thresholds in LLM generation~\citep{quach2024conformal,wang-etal-2024-conu,jiang2025conformal}. Beyond post-hoc methods, some works seek to improve calibration during training, including auxiliary calibrators or probes~\citep{zhang2025reasoning,pmlr-v235-shen24c,kossen2025semantic}, supervised fine-tuning for language models~\citep{10.5555/3737916.3740645,liu-etal-2024-llms-learn-uncertainty,li2025conftuner}, and RL objectives that jointly optimize reasoning performance and calibrated confidence~\citep{leng2025taming,damani2025binaryrewardstraininglms,liu2026cgspg}.
And though models can learn to verbalize their confidence through reinforcement learning \citep{leng2025taming, damani2025binaryrewardstraininglms, bani-harouni2026rewarding}, the quantization of these estimates to common numerical values limits their utility for fine-grained control~\citep{cruz2024evaluating}.

\section{Methodology}
We develop a calibration-aware reinforcement learning framework that jointly optimizes final-answer correctness and confidence reliability. \hyperref[subsec::empirical observation]{Subsection~\ref*{subsec::empirical observation}} motivates our use of relative calibration supervision by discussing the limitations of direct score matching in reasoning RL. \hyperref[subsec::probe training]{Subsection~\ref*{subsec::probe training}} then introduces our probe-based confidence estimator, which provides fine-grained confidence predictions at intermediate reasoning states. Finally, \hyperref[subsec::RL]{Subsection~\ref*{subsec::RL}} defines our margin-based reward over intermediate reasoning states and combines it with the answer correctness to derive the final RL training objective.

\subsection{The Cost of Calibrated Reasoning}
\label{subsec::empirical observation}
\begin{wrapfigure}[18]{r}{0.5\textwidth}
    \vspace{-1.3em}
    \centering
    \includegraphics[width=\linewidth]{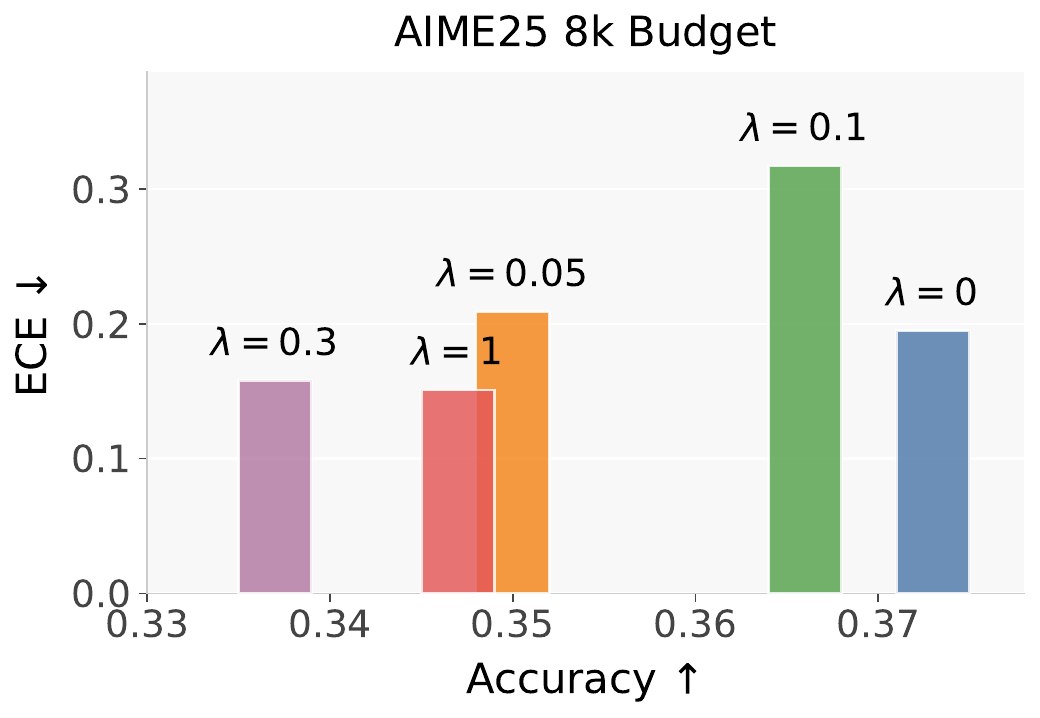}

    \caption{\small Trade-off between accuracy and expected calibration error (ECE) across Brier reward weights; $\lambda$ = 0 is the GRPO fallback.}
    \label{fig:figure3}
\end{wrapfigure}

Calibration measures how well a model's confidence aligns with its empirical accuracy. A model is perfectly calibrated if, for any confidence level $p$, the probability that its prediction is correct equals $p$, which can be written as:
$$P(\hat{Y}=Y \mid \mathrm{confidence}=p)=p.$$

One natural approach is to design a reward function that encourages the expressed confidence to match empirical accuracy via reinforcement learning~\citep{damani2025binaryrewardstraininglms}:
$$
r = \mathbb{I}(Y=\hat{Y}) - \lambda (\mathbb{I}(Y=\hat{Y})-p)^2
$$.
However, direct score matching is difficult to combine with reasoning optimization: post-training is still dominated by final-answer accuracy, while pointwise calibration penalties impose an additional per-instance target that can be noisy and only weakly aligned with the RL update direction. Our results in \autoref{fig:figure3} reflect this tension: varying the Brier reward weight moves the model along an accuracy--calibration trade-off, but does not produce a consistent Pareto improvement. This suggests that direct score matching couples calibration supervision to reasoning optimization in a brittle way, potentially degrading learning dynamics. We therefore adopt a weaker, relative form of supervision that treats calibration as a ranking problem over intermediate reasoning states (i.e., reasoning prefixes): instead of forcing confidence to match correctness at each prefix, we encourage the model to assign higher confidence to more solvable prefixes than to less solvable ones. Motivated by prior works that connect ranking objectives to calibration quality~\citep{jiang-etal-2024-addressing,wang2025always}, we instantiate this idea with a margin-based calibration objective that enlarges the confidence gap between better and worse intermediate states within the same trajectory.

\subsection{Probing Confidence in the Process}
\label{subsec::probe training}
Prior work shows that LLM hidden states encode useful signals about the model's internal confidence, which can be extracted with lightweight probes~\citep{ch-wang-etal-2024-androids,wu-etal-2025-thought,zhang2025reasoning}. Following this line of work, we attach a confidence probe to intermediate reasoning states and use it to estimate the probability that a truncated reasoning prefix will yield the correct final answer.

For a prompt $x$, let $y$ be a sampled reasoning trajectory. We select a set of compute budgets $\mathcal{B}(y)$ as predefined truncation points along the reasoning trajectory. For each budget $b \in \mathcal{B}(y)$, let $h_b(y)$ denote the policy model's final layer hidden state at the end of the truncated prefix.

To construct a supervision target for each truncated state, we append the end-of-thinking token \texttt{</think>} and force the model to generate a final-answer~\citep{muennighoff-etal-2025-s1,jurayj-etal-2025-final} (see Appendix~\ref{app:forced_answer_sampling}). Because any single forced completion can be noisy, we sample $K$ completions and use their Monte Carlo accuracy as an estimate of intermediate correctness:
\[
Y_b(y) \;=\; \frac{1}{K}\sum_{k=1}^K \mathbf{1}\!\left[\hat{a}_{b,k} = a^\star\right] \in [0,1],
\]
where $a^\star$ is the gold answer and $\hat{a}_{b,k}$ is the $k$-th forced completion from budget $b$.

The confidence probe outputs
\[
C_b(y) \;=\; \text{Probe}(h_b(y)) \in (0,1),
\]
where the probe maps the policy hidden states into probabilities. We train the probe using binary cross-entropy with soft Monte Carlo targets:
\[
\mathcal{L}_{\mathrm{probe}}
=
-\sum_{y}\sum_{b\in\mathcal{B}(y)}
\Bigl[
Y_b(y)\log C_b(y)
+
\bigl(1-Y_b(y)\bigr)\log\bigl(1-C_b(y)\bigr)
\Bigr].
\]
The probe is updated jointly with policy training so that it stays matched to the evolving rollout distribution. Its gradients are only used to update itself and do not backpropagate into the policy model. In practice, we choose the probe to be a simple 2-layer MLP.

\subsection{Reinforcement Learning with Confidence Margin}
\label{subsec::RL}

Using the intermediate correctness estimates $Y_b(y)$ and probe predictions $C_b(y)$ defined in \autoref{subsec::probe training}, we instantiate calibration supervision as a relative objective over prefixes within the same reasoning trajectory. Rather than requiring confidence to match correctness at each intermediate state, we encourage prefixes with higher estimated solvability to receive higher confidence than less solvable ones.

For each budget $b \in \mathcal{B}(y)$, let $Y_b(y)\in[0,1]$ denote the Monte Carlo estimate of intermediate correctness. We partition the selected budgets into
\[
\mathcal{B}^+(y)=\{b\in\mathcal{B}(y): Y_b(y)\geq \tfrac{1}{2}\},
\qquad
\mathcal{B}^-(y)=\{b\in\mathcal{B}(y): Y_b(y)< \tfrac{1}{2}\}.
\]
That is, $\mathcal{B}^+(y)$ contains prefixes that are more likely to lead to the correct final answer, while $\mathcal{B}^-(y)$ contains less promising ones.

We then define the margin reward as the gap between their average predicted confidence:
\[
R_{\mathrm{margin}}(y)
=
\frac{1}{|\mathcal{B}^+(y)|}\sum_{b\in\mathcal{B}^+(y)} C_b(y)
-
\frac{1}{|\mathcal{B}^-(y)|}\sum_{b\in\mathcal{B}^-(y)} C_b(y),
\]
with empty-set terms defined as $0$.
 
This objective encourages the probe to assign higher confidence to prefixes with higher estimated correctness than to those with lower estimated correctness. Because it depends only on relative confidence ordering, it is less sensitive to global confidence-scale mismatch than pointwise score-matching objectives such as the Brier penalty. Moreover, because the reward is applied to multiple intermediate budgets along each trajectory, it provides process-level supervision over the reasoning trace rather than supervision only on the final answer~\citep{lightman2024lets,luo2025improve,setlur2025rewarding}.

\textbf{Overall Reward.}
Our final reward combines answer correctness with the auxiliary margin objective:
\[
R(y)=R_{\mathrm{ans}}(y)+\lambda R_{\mathrm{margin}}(y),
\]
where $R_{\mathrm{ans}}(y)\in\{-1,1\}$ denotes the correctness reward of the full reasoning trajectory and $\lambda$ controls the strength of the auxiliary reward.
\newpage

\section{Experiments}

\subsection{Experimental Settings} 

\textbf{Baselines.}
We compare against representative baselines spanning outcome-only RL and calibration-aware RL: 
(1) \textbf{Base}~\citep{deepseekai2025deepseekr1incentivizingreasoningcapability}, the original R1-distilled Qwen-7B model before any additional RL training;
(2) \textbf{GRPO}~\citep{shao2024deepseekmathpushinglimitsmathematical}, our primary outcome-only RL baseline;
(3) \textbf{RLCR}~\citep{damani2025binaryrewardstraininglms}, a calibration-aware RL baseline that augments final-answer correctness with a Brier-style reward on verbalized confidence; and
(4) \textbf{C$^2$GSPG}~\citep{liu2026cgspg}, which incorporates confidence regularization into policy optimization.
All methods are initialized from the same base model, R1-distilled Qwen-7B and trained
with identical data, hyperparameters, and probe-training protocol (\autoref{app:impl}).

\textbf{Datasets.}
All methods are trained on the GRPO-LEAD dataset~\citep{zhang-zuo-2025-grpo}, which was initially curated to train 7B/14B models. We evaluate on a suite of mathematical reasoning benchmarks spanning a broad range of difficulty, including MATH-500~\citep{lightman2024lets}, AMC~\citep{hendrycks2021measuring}, OlympiadBench~\citep{he-etal-2024-olympiadbench}, and AIME 2024/2025. To assess out-of-domain generalization beyond mathematics, we further evaluate on scientific question answering, GPQA Diamond~\citep{rein2024gpqa}, logical reasoning, LogiQA~\citep{10.5555/3491440.3491941}, and code reasoning, LiveCodeBench~\citep{jain2025livecodebench}.

\textbf{Evaluation Metrics.} We evaluate models along three complementary dimensions: reasoning performance, calibration quality, and confidence discrimination. For reasoning performance, we report the averaged rollout accuracy (Pass@1). For calibration, we report Expected Calibration Error (ECE)~\citep{Pakdaman_Naeini_Cooper_Hauskrecht_2015}, which measures the discrepancy between predicted confidence and empirical accuracy via binning. To directly characterize overconfidence, we additionally report Positive Calibration Error (PCE)~\citep{ma2026decouplingreasoningconfidenceresurrecting}, which restricts ECE to bins where confidence exceeds accuracy:
\[
\mathrm{ECE}=\sum_{b=1}^{B}\frac{|S_b|}{n}\left|\mathrm{acc}(S_b)-\mathrm{conf}(S_b)\right|,
\quad
\mathrm{PCE}=\sum_{c(B_m)>a(B_m)}\frac{|B_m|}{N}\left|a(B_m)-c(B_m)\right|.
\]
\begin{table*}[ht]
\centering
\small
\setlength{\tabcolsep}{4.5pt}
\renewcommand{\arraystretch}{1.12}

% -------- top half --------
\begin{adjustbox}{max width=\textwidth}
\begin{tabular}{ll*{9}{c}}
\toprule
\multirow{2}{*}{Method} & \multirow{2}{*}{Conf.}
& \multicolumn{3}{c}{MATH-500}
& \multicolumn{3}{c}{AIME24}
& \multicolumn{3}{c}{AIME25} \\
\cmidrule(lr){3-5}
\cmidrule(lr){6-8}
\cmidrule(lr){9-11}
& & Acc$\uparrow$ & PCE$\downarrow$ & ECE$\downarrow$
  & Acc$\uparrow$ & PCE$\downarrow$ & ECE$\downarrow$
  & Acc$\uparrow$ & PCE$\downarrow$ & ECE$\downarrow$ \\
\midrule
Base      & probe  & .866 & \best{.000} & .132 & .398 & \second{.089} & \second{.095} & .293 & \second{.155} & \second{.155} \\
GRPO      & probe  & \second{.891} & \best{.000} & .203 & \second{.479} & .108 & .108 & \best{.373} & .195 & .195 \\
RLCR      & verbal & .888 & .065 & \best{.065} & .473 & .243 & .243 & .358 & .321 & .321 \\
C$^2$GSPG & logits & \best{.898} & \second{.003} & \second{.087} & .466 & .334 & .334 & .359 & .427 & .427 \\
\rowcolor{rlcmshade}
\textbf{RLCM} & probe  & \second{.891} & \best{.000} & .133 & \best{.492} & \best{.049} & \best{.061} & \second{.360} & \best{.120} & \best{.123} \\
\bottomrule
\end{tabular}
\end{adjustbox}

\vspace{.6em}

% -------- bottom half --------
\begin{adjustbox}{max width=\textwidth}
\begin{tabular}{ll*{9}{c}}
\toprule
\multirow{2}{*}{Method} & \multirow{2}{*}{Conf.}
& \multicolumn{3}{c}{AMC}
& \multicolumn{3}{c}{OlympiadBench}
& \multicolumn{3}{c}{\textbf{Overall}} \\
\cmidrule(lr){3-5}
\cmidrule(lr){6-8}
\cmidrule(lr){9-11}
& & Acc$\uparrow$ & PCE$\downarrow$ & ECE$\downarrow$
  & Acc$\uparrow$ & PCE$\downarrow$ & ECE$\downarrow$
  & Acc$\uparrow$ & PCE$\downarrow$ & ECE$\downarrow$ \\
\midrule
Base      & probe  & .669 & .055 & \second{.067} & .490 & .031 & \best{.059} & .543 & .066 & \second{.101} \\
GRPO      & probe  & \best{.798} & \best{.008} & \best{.045} & .566 & \second{.014} & .096 & \best{.621} & \second{.065} & .130 \\
RLCR      & verbal & .764 & .111 & .111 & \second{.568} & .205 & .205 & .610 & .189 & .189 \\
C$^2$GSPG & logits & \second{.797} & .059 & .075 & \best{.570} & .233 & .235 & \second{.618} & .211 & .232 \\
\rowcolor{rlcmshade}
\textbf{RLCM} & probe  & .785 & \second{.010} & .070 & .560 & \best{.002} & \second{.068} & \second{.618} & \best{.036} & \best{.091} \\
\bottomrule
\end{tabular}
\end{adjustbox}

\vspace{.6em}

% -------- OOD table --------
\begin{adjustbox}{max width=\textwidth}
\begin{tabular}{ll*{9}{c}}
\toprule
\multirow{2}{*}{Method} & \multirow{2}{*}{Conf.}
& \multicolumn{3}{c}{LiveCodeBench}
& \multicolumn{3}{c}{LogiQA}
& \multicolumn{3}{c}{GPQA Diamond} \\
\cmidrule(lr){3-5}
\cmidrule(lr){6-8}
\cmidrule(lr){9-11}
& & Acc$\uparrow$ & PCE$\downarrow$ & ECE$\downarrow$
  & Acc$\uparrow$ & PCE$\downarrow$ & ECE$\downarrow$
  & Acc$\uparrow$ & PCE$\downarrow$ & ECE$\downarrow$ \\
\midrule
Base      & probe  & .326 & .182 & .182 & .478 & .336 & .336 & .448 & .132 & .177 \\
GRPO      & probe  & \second{.380} & \best{.004} & \second{.130} & .490 & \second{.147} & \second{.154} & \second{.476} & \best{.021} & \second{.110} \\
RLCR      & verbal & .361 & .218 & .218 & .495 & .404 & .404 & .376 & .562 & .562 \\
C$^2$GSPG & logits & .374 & .499 & .499 & \best{.505} & .315 & .315 & .326 & .471 & .471 \\
\rowcolor{rlcmshade}
\textbf{RLCM} & probe  & \best{.389} & \second{.050} & \best{.051} & \second{.497} & \best{.127} & \best{.145} & \best{.479} & \second{.073} & \best{.102} \\
\bottomrule
\end{tabular}
\end{adjustbox}

\caption{Performance comparison on in-domain and out-of-domain benchmarks. The top two panels report results on in-domain math benchmarks, with ``Overall'' denoting the average over these tasks; the bottom panel reports results on out-of-domain tasks. ``Conf.'' denotes the confidence estimation mechanism used by each method. \textbf{Bold} and \underline{underlined} entries mark the best and second-best results, respectively. Cases where PCE equals ECE indicate that the calibration error arises entirely from overconfidence.}
\label{tab:main_results}
\end{table*}

Here, \(B\) is the number of bins, \(S_b\) the samples in bin \(b\), and \(n\) the total number of samples.

\subsection{Overall Results}
\begin{figure}[ht]
    \centering
    \small
    \includegraphics[width=0.9\linewidth]{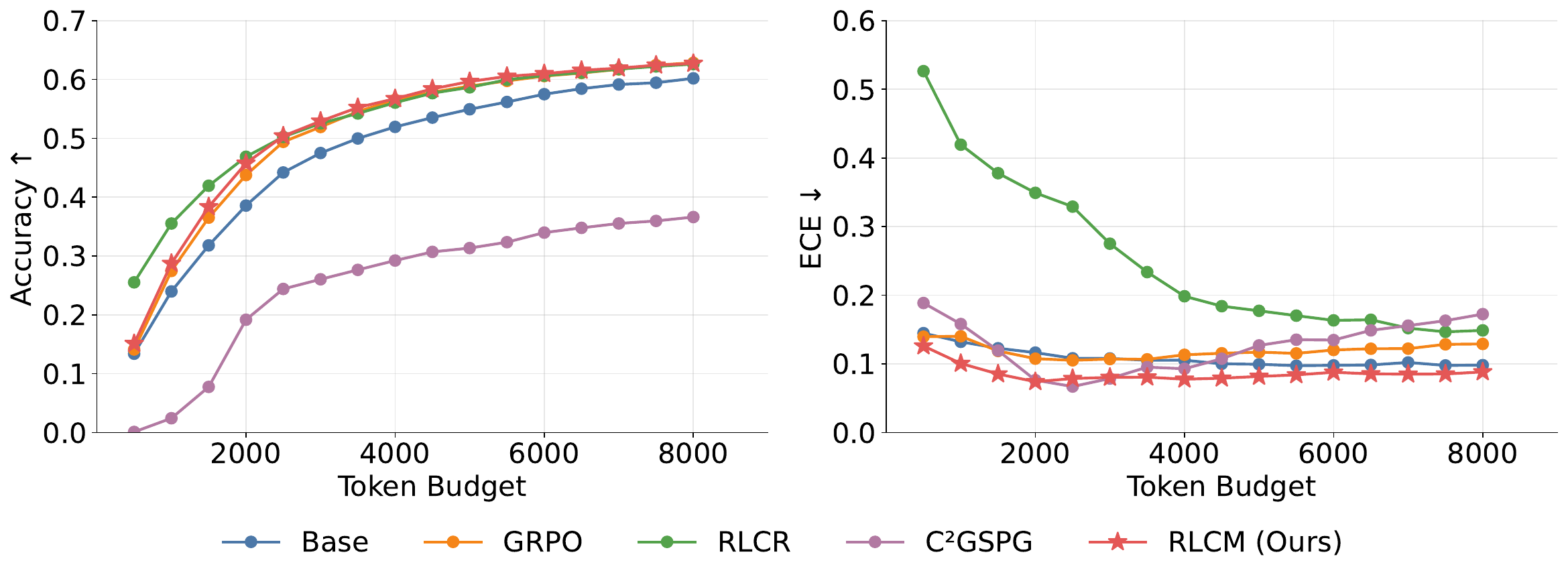}
    \caption{Average accuracy (left) and ECE (right) over five in-domain datasets across token budgets, with correctness and confidence extracted via force output. RLCM matches the strongest baseline in accuracy while consistently yielding the lowest ECE.}
    \label{fig:process_acc_ece}
\end{figure}

\textbf{Margin-based process supervision yields the best balance.}
RLCM achieves the most favorable overall trade-off between reasoning performance and calibration. As is shown in \autoref{fig:process_acc_ece}, compared to GRPO, it preserves nearly all of the accuracy gains from RL while consistently reducing calibration error. Compared to other calibration-aware methods, RLCM achieves the best accuracy and calibration on AIME24 and AIME25 while remaining competitive on the rest. These results suggest that relative supervision via a confidence-margin reward aligns better with the optimization dynamics of reasoning.

\textbf{Outcome-driven RL improves accuracy, but not calibration.}
\autoref{tab:main_results} shows that GRPO is a decent reasoning baseline with the highest overall accuracy. However, these gains do not reliably translate into better calibration. In particular, GRPO's overall calibration result is worse than the base model. On difficult benchmarks (AIME24/25), its accuracy gains are accompanied by substantial degradation in PCE and ECE. This supports our motivation: optimizing only for final-answer correctness can make the model's confidence less reliable.

\textbf{Verbalized and logit-based confidence are insufficient.}
 Although RLCR (verbalized confidence) and C$^2$GSPG (logit-based confidence) are both designed to improve calibration, they underperform GRPO not only in accuracy but also in PCE and ECE. In~\autoref{app:confidence_distributions}, their confidence signals occupy a much narrower range than RLCM's probe-based confidence, making them less expressive and more poorly calibrated. This limited dynamic range reduces their usefulness for fine-grained calibration and confidence-based control.

\textbf{The gains persist beyond the training domain.}
Although all methods are trained on math-centric data, RLCM remains strong on coding, science, and logical reasoning tasks. It achieves the best accuracy on LiveCodeBench and GPQA while also attaining the best calibration on those datasets. On LogiQA, RLCM attains the lowest calibration error while maintaining decent accuracy. These results suggest that margin-based process supervision helps preserve both performance and calibration under severe distribution shift.

\subsection{Calibrated Reasoning Process}
To verify that our method performs well throughout the reasoning trajectory rather than only at the final answer, we collect forced-output answers and probe intermediate reasoning states for confidence estimation at varying compute budgets. \autoref{fig:process_acc_ece} reports accuracy and ECE averaged across five in-domain datasets. As shown in Figure~\ref{fig:process_acc_ece} (left), RLCM matches the accuracy of the strongest baseline across all token budgets. On the calibration side (Figure~\ref{fig:process_acc_ece}, right), RLCM consistently achieves the lowest ECE across nearly the entire budget range. These results demonstrate that RLCM is the only method that delivers both high accuracy and reliable confidence estimates at every stage of reasoning, confirming that process supervision via margin reward preserves calibration throughout the trajectory rather than emerging only at the end.

\subsection{Epistemic Uncertainty Analysis}

\begin{wrapfigure}[14]{r}{0.47\textwidth}
    \vspace{-0.9em}
    \centering
    \includegraphics[width=\linewidth]{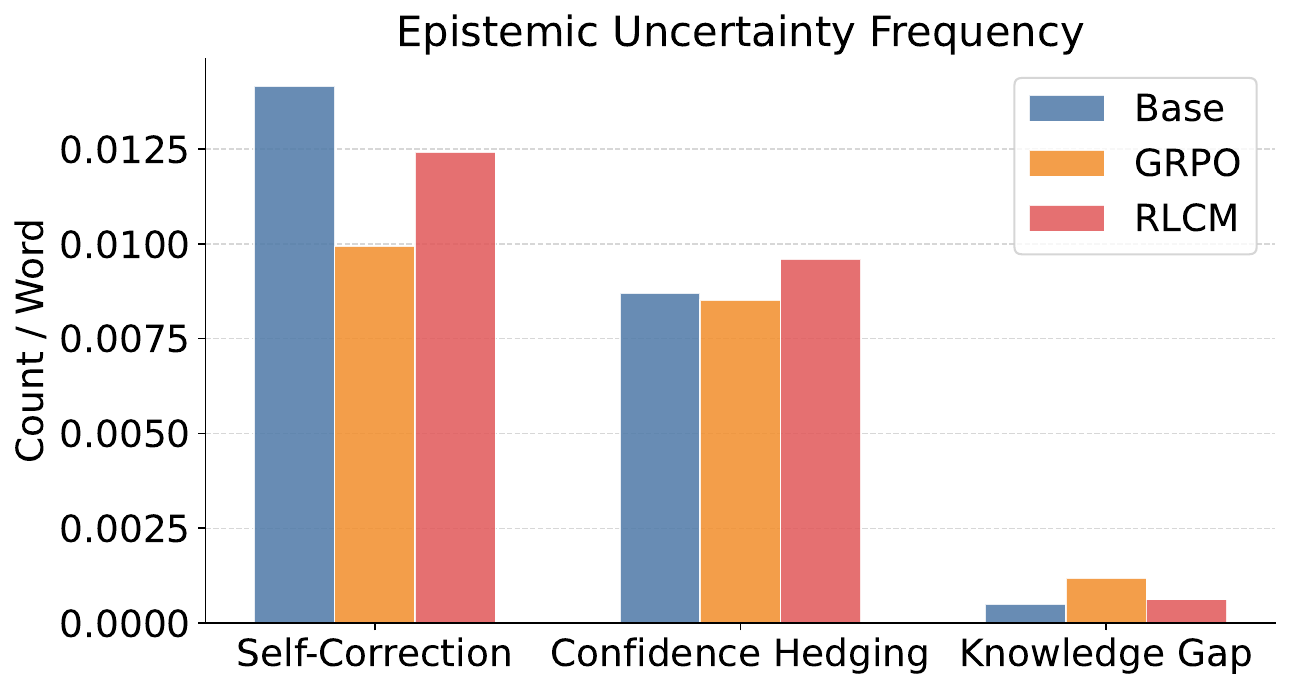}
    \caption{\small{Frequency of three types of epistemic uncertainty markers in AIME25 reasoning traces, normalized by token count.}}
    \label{fig:uncertainty_indicator}
\end{wrapfigure}
Following~\citet{qian2026demystifying, kim2026understandingreasoningllmsstrategic}, we analyze verbalized uncertainty expressions in reasoning traces to characterize model behavior at the semantic level. We group these expressions into three categories: \textit{self-correction}, \textit{confidence hedging}, and \textit{knowledge gaps}. As a case study on AIME25 (\autoref{fig:uncertainty_indicator}), RLCM largely preserves the distribution of uncertainty expressions exhibited by the base model across all three categories, whereas GRPO markedly suppresses self-correction behavior. Notably, this preservation emerges despite the absence of any explicit reward on semantic-level uncertainty, suggesting that token-level calibration naturally aligns with the base model's verbalized reasoning patterns. More analysis can be found in the \autoref{sec::semantic_uncertainty}.

\subsection{Ablation Study}
This ablation isolates two design choices: where calibration supervision is applied (final or intermediate) and how it is applied (Brier score or margin reward). For a fair comparison, all variants use the same on-the-fly probe-based confidence estimator and the same auxiliary reward weight, $\lambda=0.1$. Thus, they are closely matched variants of our proposed framework, RLCM. The only difference is in the form and placement of the calibration signal.

\begin{table*}[ht]
\centering
\small
\setlength{\tabcolsep}{4.5pt}
\renewcommand{\arraystretch}{1.12}

\begin{adjustbox}{max width=\textwidth}
\begin{tabular}{l*{9}{c}}
\toprule
\multirow{2}{*}{Method}
& \multicolumn{3}{c}{OlympiadBench}
& \multicolumn{3}{c}{AIME24}
& \multicolumn{3}{c}{AIME25} \\
\cmidrule(lr){2-4}
\cmidrule(lr){5-7}
\cmidrule(lr){8-10}
& Acc$\uparrow$ & PCE$\downarrow$ & ECE$\downarrow$
& Acc$\uparrow$ & PCE$\downarrow$ & ECE$\downarrow$
& Acc$\uparrow$ & PCE$\downarrow$ & ECE$\downarrow$ \\
\midrule
Final-Brier   & \best{.564} & .200 & .200
              & \best{.496} & .224 & .224
              & \best{.366} & .317 & .317 \\
Final-Margin  & .526 & .126 & .126
              & .456 & .119 & .122
              & .328 & .201 & .201 \\
Process-Brier & .549 & \second{.100} & \second{.100}
              & \second{.493} & \second{.061} & \second{.064}
              & .350 & \second{.171} & \second{.171} \\
\rowcolor{rlcmshade}
\textbf{RLCM} & \second{.560} & \best{.002} & \best{.068}
              & .492 & \best{.049} & \best{.061}
              & \second{.360} & \best{.120} & \best{.123} \\
\bottomrule
\end{tabular}
\end{adjustbox}

\caption{
Effects of final and process reward variants. Darker shaded cells with bold text indicate the best result in each column, and lighter shaded cells indicate the second-best result. The RLCM row is lightly shaded for readability.
}
\label{tab:ablation_study}
\end{table*}

We compare four variants: (1) \textbf{Final-Brier}, which applies a Brier-style penalty only to the final-step probe confidence, $R(y)=R_{\mathrm{ans}}(y)-\lambda\bigl(Y(y)-C(y)\bigr)^2$;
(2) \textbf{Final-Margin}, the final-step counterpart of our margin reward, $R(y)=R_{\mathrm{ans}}(y)+\lambda\,\mathbb{I}(Y=\hat{Y})\,C(y)$;
(3) \textbf{Process-Brier}, which aligns intermediate probe confidence with intermediate Monte Carlo correctness using an averaged Brier penalty,
$R(y)=R_{\mathrm{ans}}(y)-\lambda\frac{1}{|\mathcal{B}(y)|}\sum_{b\in\mathcal{B}(y)}\bigl(Y_b(y)-C_b(y)\bigr)^2$;
and (4) \textbf{RLCM}, which applies margin-based process supervision over intermediate completions.

The ablation study in \autoref{tab:ablation_study} highlights two main findings. First, process supervision consistently outperforms final step only supervision, improving calibration while largely preserving reasoning accuracy. Second, margin-based supervision produces more reliable calibration than Brier-style score matching, especially in reducing overconfidence. In contrast, \textbf{Final-Brier} performs poorly: its ECE and PCE are nearly identical across benchmarks, suggesting that almost all of its calibration error arises from overconfidence. Overall, these results indicate that process-level margin supervision provides a more stable and effective calibration signal than direct Brier matching.
Among all variants, \textbf{RLCM} provides the strongest overall balance between accuracy and calibration.

See ablation study on the base model, hyperparameter $\lambda$ in the \autoref{app:ablation}.

\section{Downstream Applications}

\subsection{Calibrated Reasoning for Risk Control}
Conformal Risk Control~\citep{angelopoulos2024conformal} extends conformal prediction~\citep{10.5555/1390681.1390693} to control the expected value of a monotone risk function. We seek a parameter $\lambda$ such that the risk on unseen data is bounded by $\alpha$ with probability at least $1-\delta$. We solve this using Learn-Then-Test (LTT)~\citep{10.1214/24-AOAS1998}, which frames threshold selection as a multiple-testing problem over hypotheses whose null states that the risk exceeds $\alpha$. By combining sequential testing with concentration bounds, LTT returns a statistically valid threshold. 
However, validity does not imply efficiency: when confidence is uncalibrated or non-monotonic, LTT tends to select conservative thresholds, reducing the compute savings of dynamic inference. Following \citet{wang2026conformalthinkingriskcontrol}, we therefore use two confidence thresholds, $\lambda_1$ and $\lambda_2$. $\lambda_1$ triggers early exit under persistently low confidence to save computation on likely unsolved examples, while $\lambda_2$ triggers exits once confidence is high enough to justify a reliable prediction.

\begin{figure}[ht]
    \centering
    \tiny
    \includegraphics[width=0.9\linewidth]{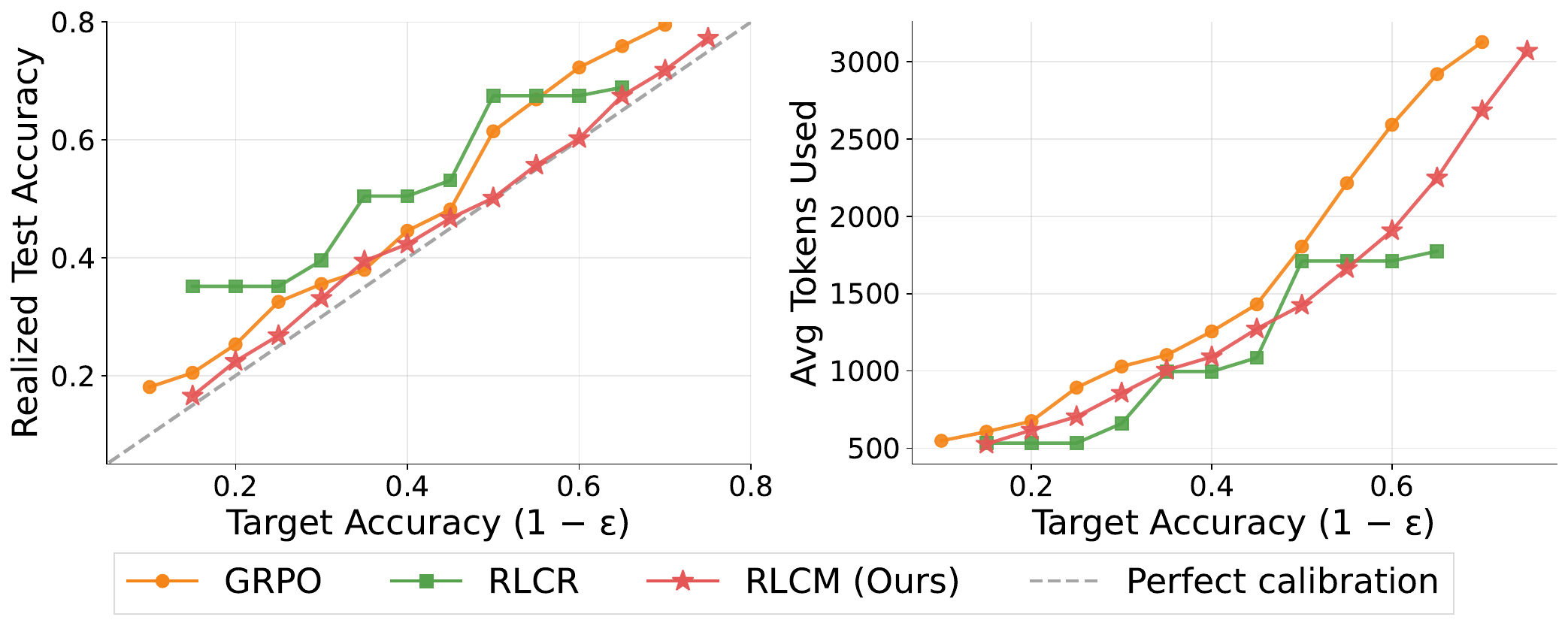}
    \caption{\textbf{Risk control on AMC with LTT.} Compared with GRPO and RLCR, RLCM yields a smooth, calibrated target-to-realized accuracy curve (Left), uses fewer tokens at comparable target accuracies (Right), and indicates more useful confidence for early exiting.}
    \label{fig:ltt_combined_amc}
\end{figure}

\autoref{fig:ltt_combined_amc} shows that RLCM provides the most useful confidence signal for LTT-based early exit. Although all methods successfully control risk, RLCM's realized accuracy tracks the perfect-calibration line more closely than GRPO or RLCR, with a smoother curve that enables finer-grained threshold control. GRPO is consistently overly conservative, spending excess compute to achieve unnecessarily high accuracy. RLCR, by contrast, exhibits step-like plateaus due to its coarse verbalized confidence, which limits precise risk control.
This advantage also translates into compute savings. At comparable target accuracies, RLCM uses substantially fewer tokens than GRPO and offers a broader, smoother accuracy--compute trade-off than RLCR, indicating that its probe-based confidence is more effective, and more efficient for continuous early-exit control.

\subsection{Confidence-Weighted Aggregation}

Well-calibrated confidence can improve answer aggregation by weighting candidate answers according to their estimated reliability rather than counting all rollouts equally. Given $N$ sampled rollouts producing answers $\{a_i\}_{i=1}^{N}$ with confidence scores $\{c_i\}_{i=1}^{N}$, and letting $\mathcal{A}$ denote the set of distinct answers, majority voting selects the most frequent answer, while confidence-weighted voting selects the answer with the highest total confidence:

\centerline{$\hat{a}_{\mathrm{Maj}}
=
\arg\max_{a\in\mathcal{A}} \left| \{ i : a_i = a \} \right|,
\qquad
\hat{a}_{\mathrm{Conf}}
=
\arg\max_{a\in\mathcal{A}} \sum_{i:\, a_i = a} c_i .
$}

Here ties in majority voting are broken uniformly at random, and $c_i$ denotes the confidence assigned to rollout $i$. Pass@1 denotes the expected accuracy of a single sampled rollout.

\begin{figure}[htbp]
    \centering
    \small
    \includegraphics[width=\linewidth]{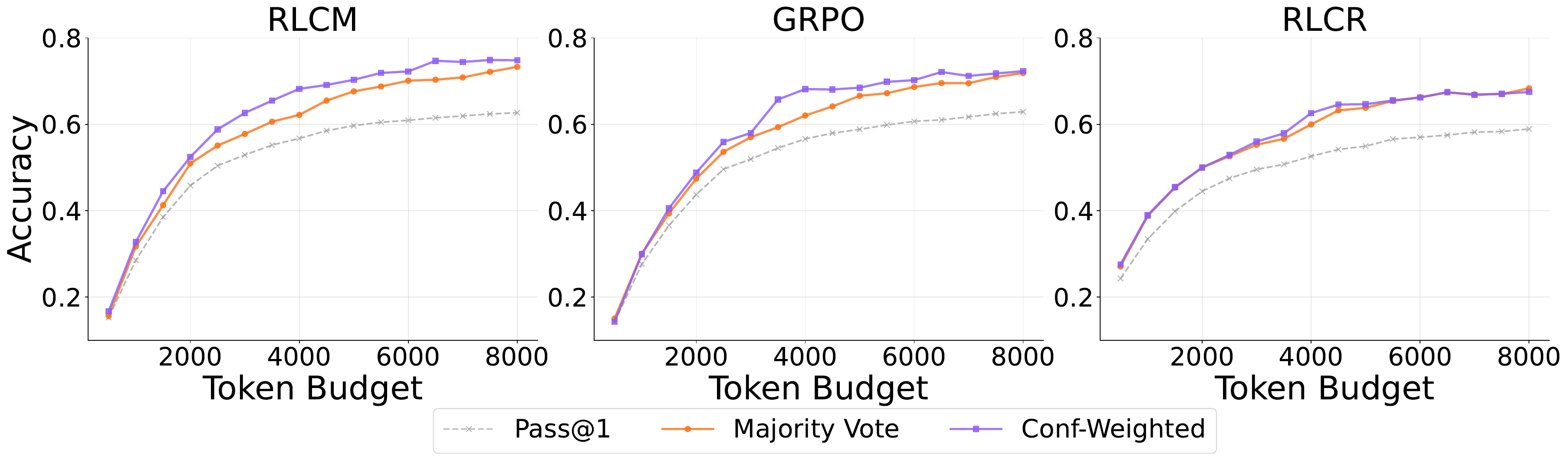}
    \caption{RLCM's margin reward encourages it to expose useful uncertainty scores to the probe at every reasoning step, yielding the strongest confidence-weighted aggregation results compared to GRPO and RLCR}
    \label{fig:conf-weighted-voting}
\end{figure}

\begin{table}[htbp]

\centering
\small
\setlength{\tabcolsep}{5pt}
\begin{tabular}{l*{9}{c}}
\toprule
\multirow{2}{*}{Method}
& \multicolumn{3}{c}{RLCM}
& \multicolumn{3}{c}{GRPO}
& \multicolumn{3}{c}{RLCR}  \\
\cmidrule(lr){2-4}
\cmidrule(lr){5-7}
\cmidrule(lr){8-10}
& Pass@1 & Maj. & Conf. & Pass@1 & Maj. & Conf. & Pass@1 & Maj. & Conf. \\
\midrule
Accuracy & .627 & \textbf{.733} & \textbf{.748}
         & \textbf{.629} & .719 & .723
         & .589 & .683 & .675 \\
\bottomrule
\end{tabular}

\caption{Across Benchmarks Average Accuracy at 8k budget comparison under different aggregation strategies. RLCM yields substantial improvements when aggregating answers across parallel attempts.}
\label{tab:accuracy-aggregation}
\end{table}

\autoref{tab:accuracy-aggregation} and \autoref{fig:conf-weighted-voting} show that similar pass@1 accuracy does not imply equally useful confidence for aggregation. 
Although RLCM and GRPO have nearly identical Pass@1 accuracy, RLCM delivers consistently larger gains from confidence-weighted voting, indicating that its confidence estimates are better aligned with correctness. In contrast, GRPO and RLCR exhibit limited improvements, suggesting that their confidence signals are less reliable for weighting candidate answers. Detailed per-benchmark results are provided in  \autoref{sec::app_conf_weight}.

\section{Conclusion}
We presented RLCM, a calibration-aware reinforcement learning framework that uses margin-based process supervision to enable calibrated LLM reasoning. By encouraging higher confidence on more solvable intermediate states than on less reliable ones, RLCM learns more trustworthy confidence estimates without sacrificing reasoning performance. Empirically, this leads to stronger calibration across diverse benchmarks and makes confidence substantially more useful for downstream decision making, including conformal risk control and confidence-weighted aggregation. These results highlight relative process supervision as a practical and effective strategy for training reliable reasoning models.

\section*{Acknowledgments}
This work is supported by the project of ``Safely Confident Reasoning'' in the Science of Trustworthy AI program of the Schmidt Sciences, the Defense Advanced Research Projects Agency (DARPA) under Contract No. HR001125C0304, and Office of Naval Research funding ONR:N00014-24-1-2089. Any opinions, findings and conclusions or recommendations expressed in this material are those of the author(s) and
do not necessarily reflect the views of Schmidt Sciences, DARPA, or ONR.

\bibliography{colm2026_conference}
\bibliographystyle{colm2026_conference}

\newpage
\appendix
\section{Forced-Answer Sampling from Intermediate Prefixes}
\label{app:forced_answer_sampling}

This section describes how we construct the intermediate supervision target used in \autoref{subsec::probe training}. Our goal is to estimate what the model already knows at a truncated reasoning state, rather than what it could discover after continuing to think.

Given a prompt $x$ and a sampled reasoning trajectory $y$, we truncate the trajectory at budget $b$ and convert the truncated prefix into a \emph{forced-answer prompt}. Concretely, we terminate the reasoning trace with \texttt{</think>} and then immediately ask the model to provide its final answer. The template is:

\promptbox{
\textbf{User:}\\
\hspace*{1em}[original problem]\\[4pt]
\textbf{Assistant:}\\
\texttt{<think>}\\
\hspace*{1em}[partial reasoning trace up to budget $b$]\\
\texttt{</think>}\\
If I were to give the final answer now, the final answer would be \texttt{\textbackslash boxed\{}
}

This construction serves two purposes. First, it prevents the model from continuing its chain-of-thought beyond the selected budget. Second, it asks for a compact answer in a standardized format, making answer extraction and verification straightforward.

From this forced-answer prompt, we sample $K$ short continuations. Each continuation fills in the contents of the boxed answer, yielding a sampled prediction $\hat{a}_{b,k}$. We then compare each sampled answer against the gold answer $a^\star$ and compute the empirical success rate
\[
Y_b(y)
=
\frac{1}{K}\sum_{k=1}^{K}\mathbf{1}\!\left[\hat{a}_{b,k}=a^\star\right].
\]

We interpret $Y_b(y)$ as an estimate of the conditional probability that the truncated reasoning prefix already supports the correct answer. Importantly, this target measures \emph{answerability at the current budget}, not the success rate of the full untruncated trajectory.

In practice, this forced-answer sampling procedure is simple, cheap, and compatible with online RL training (where we just write a custom secondary loop that runs fast in VeRL~\citep{sheng2024hybridflow}). It produces a dense supervision signal across intermediate reasoning states while avoiding the need for token-level human annotations or expensive external judges.

\section{Implementation Details}
\label{app:impl}
All methods are trained with a batch size of $32$, learning rate $10^{-6}$, 6 rollouts,
temperature $0.8$, $8$k maximum generation tokens, and $600$ update steps
($\sim2.5$ epochs). For probe training in RLCM, we sample $K=4$ forced completions per
truncated prefix to obtain the probe's target. At evaluation time we decode with
temperature $0.6$, sampling 32 rollouts per problem for the AIME benchmarks and 4
rollouts for the remaining benchmarks, which contain substantially more problems;
all other decoding hyperparameters are held fixed across methods. All experiments are
conducted on a single 4$\times$H100 NVL node.

Since Base and GRPO do not come with a probe, we train one for each. We sample 8
rollouts over 80 random problems from the GRPO-LEAD dataset and force a completion
every 500 tokens to provide the probe a confidence target, where the probe input is
the hidden state of the token at the cut-off. This yields approximately $10{,}000$
data points, which we split $8{:}2$ into train and validation, training for 100 steps
with early stopping monitored on the validation set.

\section{Ablation Study}
\label{app:ablation}

\subsection{Change of Base Model}
\label{app:ablation-base}

Our main experiments use DeepSeek-R1-Distill-Qwen-7B, a distilled reasoning model that
already begins with strong long-form math competence. To test whether RLCM's
gains depend on this starting point, we repeat training on Qwen3-4B-Instruct-2507~\citep{yang2025qwen3technicalreport}, which
differs along two axes: model family and scale, and reasoning versus non-reasoning initialization. All methods are trained with identical data, hyperparameters, and post-hoc probe-training protocol as the main paper experiments (\autoref{app:impl}).

\autoref{tab:ablation-base} reports results on AIME24 and AIME25, the most challenging reasoning benchmarks in our setting, together with GPQA-Diamond as an out-of-domain test. RLCM attains the highest accuracy on all three benchmarks and the lowest ECE on both AIME benchmarks, reducing AIME25 ECE from $.169$ (Base) and $.132$ (GRPO) to $.056$. The out-of-domain result is more telling: on GPQA-Diamond, GRPO improves accuracy over Base but degrades calibration markedly ($.218 \rightarrow .272$), whereas RLCM improves accuracy while holding calibration at the Base level. This is the same pattern we observe on the distilled model, where outcome-only RL purchases accuracy at the cost of confidence reliability and the margin reward is what closes the gap. RLCM is therefore not specific
to distilled reasoning initialization.

\begin{table}[ht]
\centering
\small
\begin{tabular}{lcccccc}
\toprule
& \multicolumn{2}{c}{AIME24} & \multicolumn{2}{c}{AIME25} & \multicolumn{2}{c}{GPQA-Diamond} \\
\cmidrule(lr){2-3} \cmidrule(lr){4-5} \cmidrule(lr){6-7}
Method & Acc~$\uparrow$ & ECE~$\downarrow$ & Acc~$\uparrow$ & ECE~$\downarrow$ & Acc~$\uparrow$ & ECE~$\downarrow$ \\
\midrule
Base & .552 & .073 & .427 & .169 & .523 & \textbf{.218} \\
GRPO & .567 & .081 & .435 & .132 & .569 & .272 \\
RLCM & \textbf{.570} & \textbf{.065} & \textbf{.458} & \textbf{.056} & \textbf{.576} & .219 \\
\bottomrule
\end{tabular}
\caption{Accuracy and ECE with Qwen3-4B-Instruct as the base model. RLCM achieves the best accuracy on all three benchmarks and the lowest ECE on both AIME benchmarks. On the out-of-domain GPQA-Diamond, GRPO raises accuracy but degrades calibration relative to Base, while RLCM raises accuracy at no calibration cost.}
\label{tab:ablation-base}
\end{table}

\subsection{Sensitivity to the Margin Reward Weight $\lambda$}
\label{app:ablation-lambda}

Our main results fix the margin reward weight at $\lambda = 0.1$. To verify that RLCM's improvements are a property of the margin objective rather than of this particular value, we sweep $\lambda \in \{0.05, 0.1, 0.3, 0.5\}$ on DeepSeek-R1-Distill-Qwen-7B, holding all other hyperparameters fixed.

As shown in \autoref{tab:ablation-tab-lambda}, RLCM remains in a tight band across the full range: accuracy stays within $[.482, .500]$ and ECE within $[.061, .077]$, with no systematic accuracy--calibration trade-off as $\lambda$ grows. This contrasts sharply with the pointwise Brier reward (\autoref{fig:figure3}), where varying the calibration
weight moves the model along a trade-off curve without yielding a consistent Pareto improvement. We attribute the difference to the weaker constraint the margin reward imposes: it asks only that more-solvable prefixes score higher than less-solvable ones, rather than pinning each confidence to an absolute target, which leaves the policy more freedom to optimize correctness. We adopt $\lambda = 0.1$ simply as one well-balanced
point within this stable range.

\begin{table}[ht]
\centering
\small
\begin{tabular}{lcccc}
\toprule
$\lambda$ & 0.05 & 0.1 & 0.3 & 0.5 \\
\midrule
AIME24 Acc~$\uparrow$ & \textbf{.500} & .492 & .496 & .482 \\
AIME24 ECE~$\downarrow$ & .065 & \textbf{.061} & .077 & \textbf{.061} \\
\bottomrule
\end{tabular}
\caption{Sensitivity of RLCM to the margin reward weight $\lambda$ on AIME24, with all
other hyperparameters held fixed. Both accuracy and ECE remain stable across an order of
magnitude in $\lambda$, indicating that RLCM's gains are not an artifact of the
$\lambda = 0.1$.}
\label{tab:ablation-tab-lambda}
\end{table}

\section{Confidence Distribution Analysis}
\label{app:confidence_distributions}

\begin{figure}[ht]
    \centering
    \includegraphics[width=\linewidth]{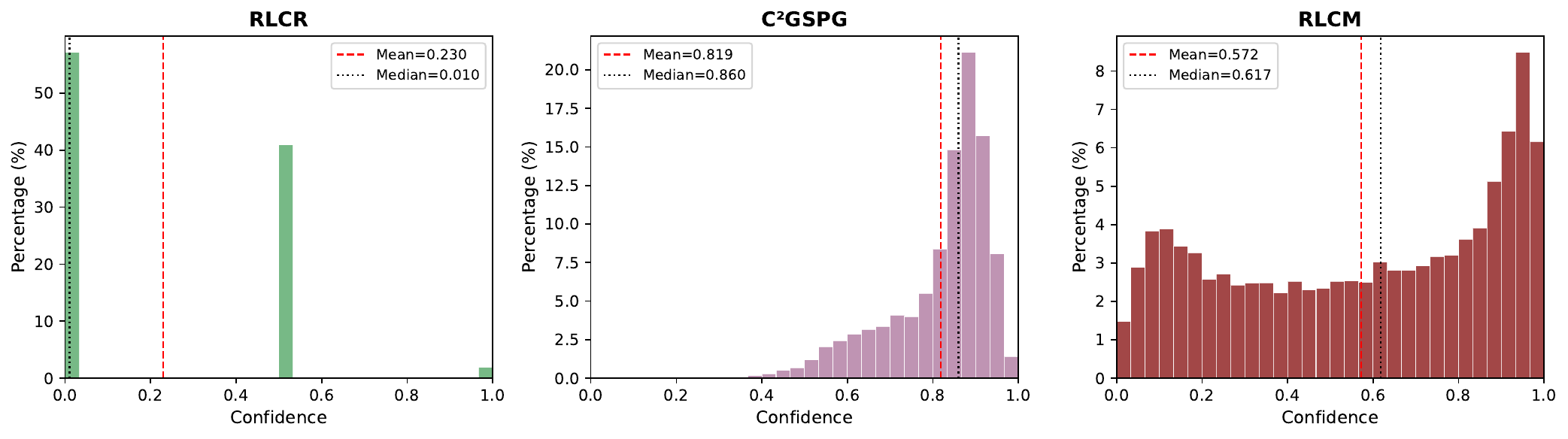}
    \caption{
    Confidence distributions for RLCR, C$^2$GSPG, and RLCM. RLCR is highly quantized, C$^2$GSPG is compressed in a narrow high-confidence range, and RLCM uses a broader portion of the confidence scale. Red dashed lines mark the mean; black dotted lines mark the median.
    }
    \label{fig:confidence_distributions}
\end{figure}
\subsection{Observation}
Figure~\ref{fig:confidence_distributions} helps explain why RLCR and C$^2$GSPG underperform despite being designed for calibration. Both produce confidence scores with limited dynamic range, but in different ways.

RLCR is highly discrete, with most mass near zero and a smaller spike around $0.5$. This suggests that verbalized confidence provides only a coarse signal rather than a smooth uncertainty estimate. Its mean is $0.230$, while its median is only $0.010$, showing that most predictions receive extremely low confidence.

C$^2$GSPG shows the opposite pattern: its scores are concentrated in a narrow high-confidence band, with mean $0.819$ and median $0.860$. This compression reduces resolution, since many examples receive similarly large confidence values regardless of difficulty.

RLCM produces a much broader and smoother distribution over $[0,1]$. Its mean and median ($0.572$ and $0.617$) indicate that it avoids both the low-confidence collapse of RLCR and the upper-tail saturation of C$^2$GSPG. This wider spread makes the confidence signal more useful for thresholding, abstention, and confidence-weighted aggregation.

Overall, RLCM does not only reduce calibration error; it also yields a more expressive confidence representation.

\subsection{Analysis}
Figure~\ref{fig:confidence_distributions} suggests that RLCR and C$^2$GSPG suffer from two different expressivity bottlenecks.

RLCR is limited by \emph{quantization}. Its confidence is produced as a verbalized scalar after the reasoning trace and final answer, so a rich internal uncertainty state must be mapped into a small set of preferred numeric outputs. In practice, recent work shows that verbalized confidence in LLMs is heavily concentrated on a few round-number anchors, and can be weakly aligned with the model's internal calibration signal, especially when reasoning and confidence are generated together \citep{dai2026rescalingconfidence,miao2026closing}. This is consistent with the highly discrete distribution in Figure~\ref{fig:confidence_distributions}: RLCR provides a coarse confidence signal, rather than a smooth estimate of uncertainty.

C$^2$GSPG exhibits the opposite failure mode: \emph{compression}. It defines confidence using normalized sequence probability,
\[
c(y_{1:T})
=
\exp\!\left(\frac{1}{T}\sum_{t=1}^T \log \pi(y_t \mid x, y_{<t})\right),
\]
which averages token-level certainty over the entire reasoning trace. However, the branching-factor analysis by \citet{yang2025probabilityconcentration} shows that aligned language models typically become more predictable later in generation, and that longer chain-of-thought pushes decoding into increasingly low-entropy stages. As a result, late high-probability tokens dominate the average, washing out earlier uncertainty and pulling confidence into a narrow high-confidence band. This explains why C$^2$GSPG has limited dynamic range even though it avoids the discrete outputs of RLCR.

RLCM avoids both bottlenecks. Instead of verbalizing confidence at the end or averaging sequence probability over the whole trace, it estimates confidence from intermediate reasoning states and trains this signal with relative process supervision. The broader distribution in Figure~\ref{fig:confidence_distributions} therefore reflects a more expressive confidence representation: it preserves meaningful variation across examples, which is more useful for thresholding, abstention, and confidence-weighted aggregation.

\section{Epistemic Uncertainty Analysis}
\label{sec::semantic_uncertainty}
We analyze verbalized uncertainty expressions in reasoning traces to characterize model behavior at the semantic level. We define three categories of uncertainty indicators: \textbf{Self-Correction} (e.g., \textit{wait}, \textit{let me reconsider}), \textbf{Confidence Hedging} (e.g., \textit{perhaps}, \textit{maybe}), and \textbf{Knowledge Gap} (e.g., \textit{I don't know}), and identify their occurrences via lexical pattern matching over the generated rollouts. The full list of phrases used for each category is given in \autoref{tab:uncertainty_expressions}.

As a case study on AIME25 (\autoref{fig:uncertainty_indicator}), RLCM largely preserves the distribution of uncertainty expressions exhibited by the base model across all three categories, whereas GRPO markedly suppresses self-correction behavior. Notably, this preservation emerges despite the absence of any explicit reward on semantic-level uncertainty, suggesting that token-level calibration naturally aligns with the base model's verbalized reasoning patterns.

\begin{figure}
    \centering
    \includegraphics[width=\linewidth]{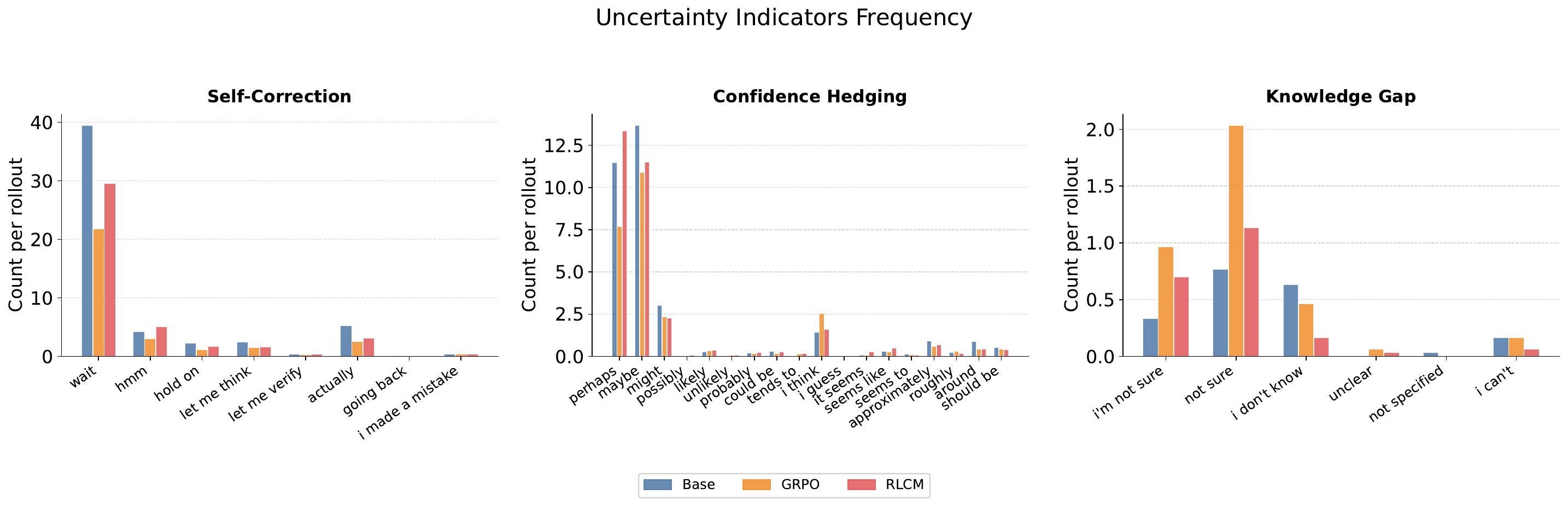}
    \caption{
Frequency of uncertainty indicators across three semantic categories, measured as average count per rollout. RLCM produces fewer hedging and knowledge-gap expressions than Base and GRPO while maintaining a similar rate of self-correction, indicating that calibration gains are reflected in reasoning content.
}
    \label{fig:detail_uncertainty_indicators}
\end{figure}

Breaking this down further, \autoref{fig:detail_uncertainty_indicators} reports the average per-rollout frequency of each individual indicator for Base, GRPO, and RLCM. RLCM produces fewer hedging and knowledge-gap expressions than both baselines while maintaining a comparable rate of self-correction. Taken together, these results suggest that RLCM's calibration gains are reflected in the substance of reasoning rather than in surface-level uncertainty cues.

\begin{table}[ht]
\centering
\small
\begin{tabular}{p{3.2cm} p{10.5cm}}
\toprule
\textbf{Category} & \textbf{Example Phrases} \\
\midrule
\textbf{Self-Correction} &
\textit{wait}, \textit{hmm}, \textit{hold on}, \textit{let me reconsider}, \textit{let me redo}, \textit{let me think}, \textit{let me verify}, \textit{let's recalculate}, \textit{actually}, \textit{on second thought}, \textit{going back}, \textit{scratch that}, \textit{I made a mistake}, \textit{that's wrong}, \textit{that was wrong} \\
\midrule
\textbf{Confidence Hedging} &
\textit{perhaps}, \textit{maybe}, \textit{might}, \textit{possibly}, \textit{likely}, \textit{unlikely}, \textit{probably}, \textit{could be}, \textit{tends to}, \textit{I think}, \textit{I believe}, \textit{I guess}, \textit{I suspect}, \textit{it seems}, \textit{seems like}, \textit{seems to}, \textit{it appears}, \textit{apparently}, \textit{presumably}, \textit{approximately}, \textit{roughly}, \textit{around}, \textit{should be}, \textit{I'm not certain}, \textit{I'm unsure} \\
\midrule
\textbf{Knowledge Gap} &
\textit{I'm not sure}, \textit{not sure}, \textit{I don't know}, \textit{unclear}, \textit{without more}, \textit{not specified}, \textit{I cannot}, \textit{I can't} \\
\bottomrule
\end{tabular}
\caption{Taxonomy of epistemic uncertainty indicators. We group surface-level uncertainty expressions into three categories: Self-Correction, Confidence Hedging, and Knowledge Gap and use the listed phrases as lexical patterns to identify regions of externalized uncertainty in reasoning traces.}
\label{tab:uncertainty_expressions}
\end{table}

\section{Confidence-Weighted Aggregation}
\label{sec::app_conf_weight}

\begin{figure}[ht]
    \centering
    \includegraphics[width=0.95\linewidth]{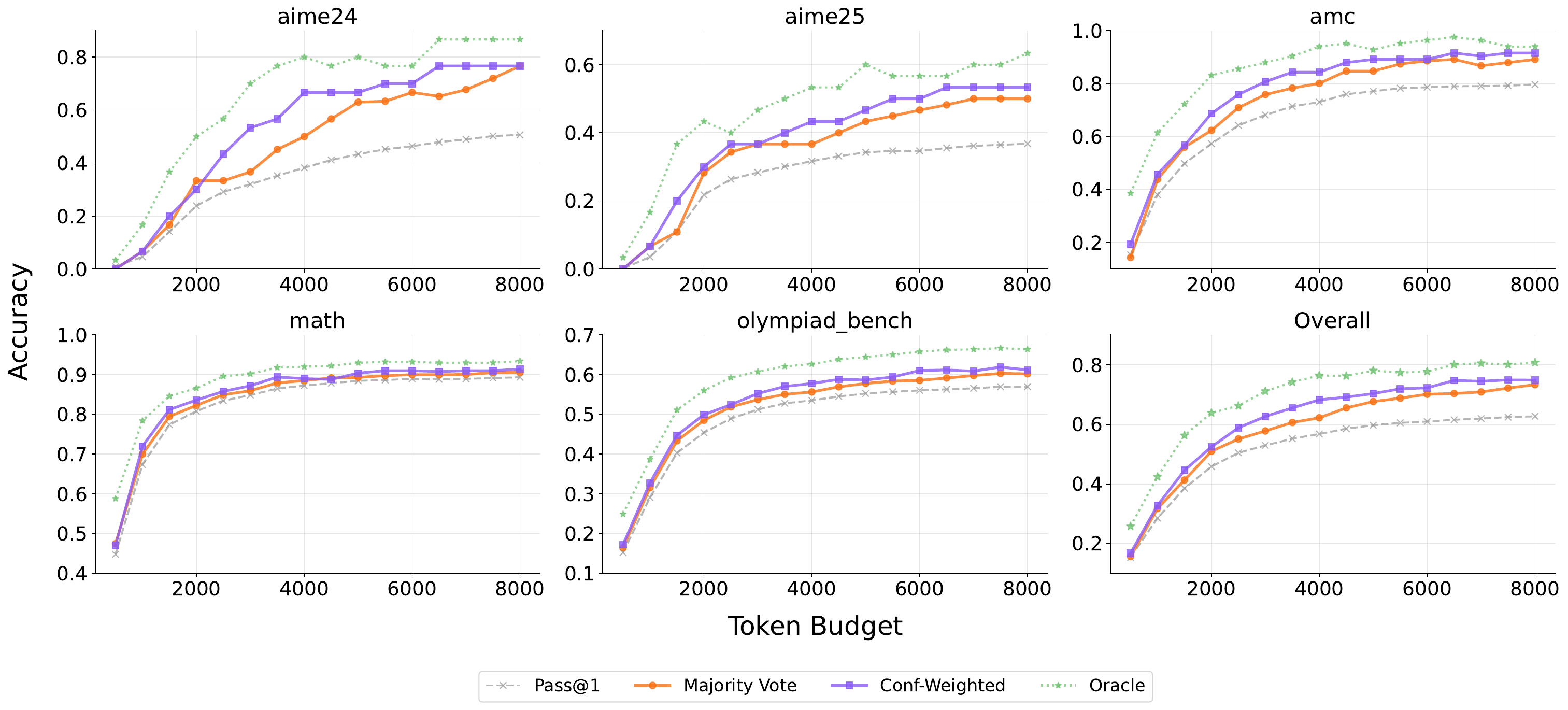}
    \includegraphics[width=0.95\linewidth]{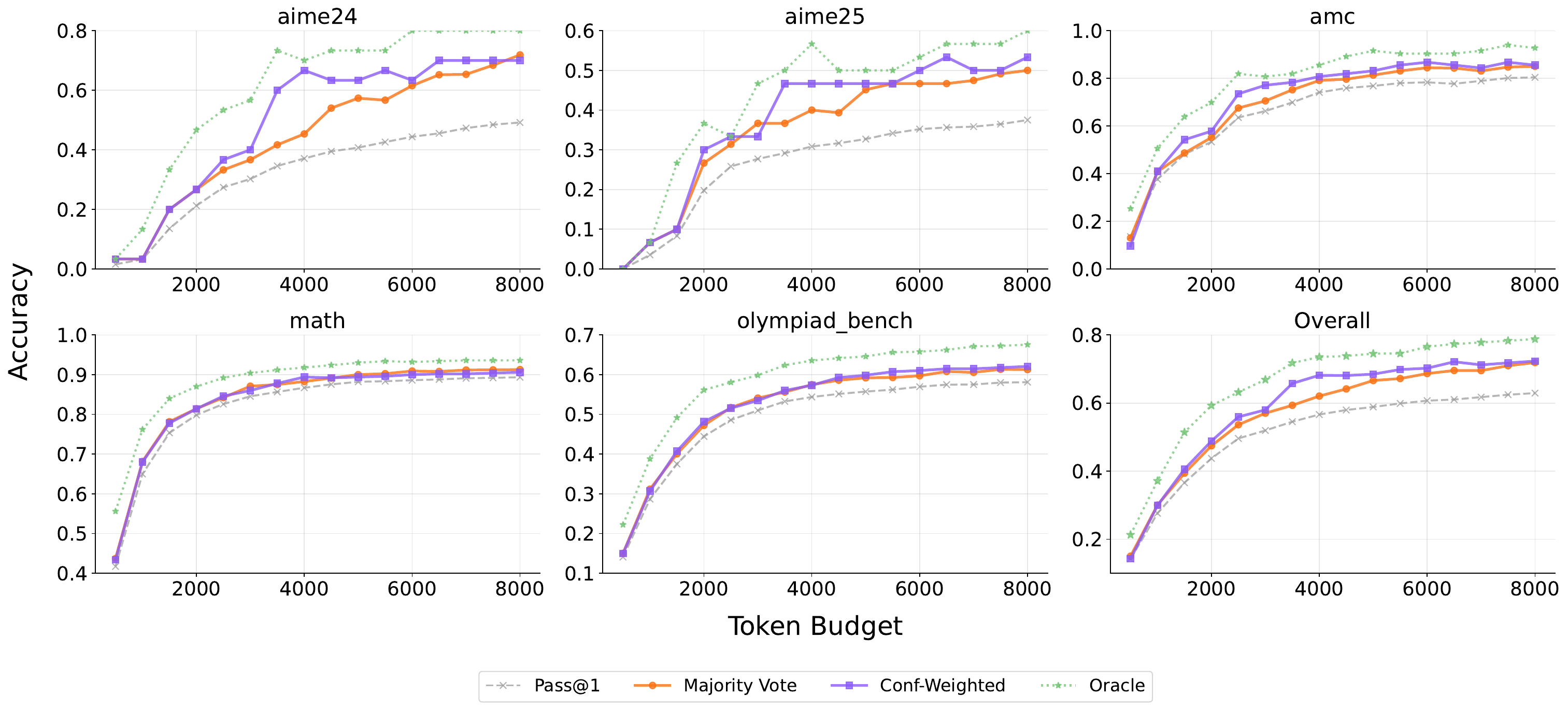}
    \includegraphics[width=0.95\linewidth]{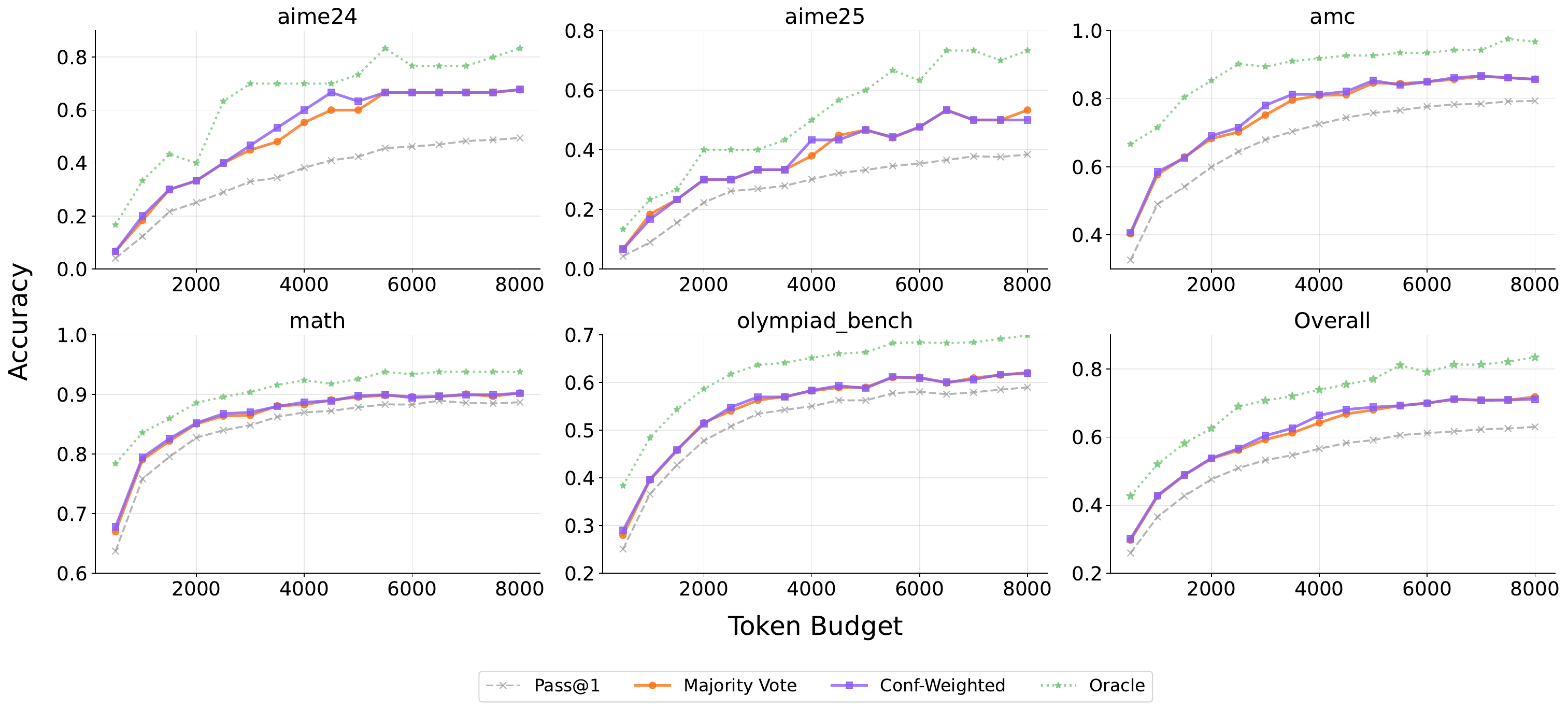}
    \caption{Per-benchmark aggregation performance across token budgets for \textsc{RLCM}, \textsc{GRPO}, and \textsc{RLCR}. Each row corresponds to one method, and each panel compares Pass@1, majority voting, confidence-weighted voting, and oracle aggregation. \textsc{RLCM} shows the most consistent gains from confidence-weighted voting over plain majority voting, indicating that its confidence estimates are more reliably aligned with answer correctness.}
    \label{fig:conf_weight_agg_per_bench}
\end{figure}

We report per-benchmark aggregation results here. In addition to Pass@1, majority voting,
and confidence-weighted voting, \autoref{fig:conf_weight_agg_per_bench} includes an \textbf{Oracle} strategy that selects a correct answer whenever one exists among the $N$ sampled rollouts. Oracle upper-bounds what any such rule can achieve over the same candidate set, so the gap to it separates accuracy lost to imperfect answer selection from accuracy lost to reasoning
failure.

Moving from majority to confidence-weighted voting changes average accuracy by $+.015$ for
RLCM, $+.004$ for GRPO, and $-.008$ for RLCR (\autoref{tab:accuracy-aggregation}). The sign
flip for RLCR reflects its highly quantized verbalized confidence
(\autoref{fig:confidence_distributions}), which supplies weights that are close to uninformative. This is the distinction the margin reward targets: ECE is a marginal statistic over the evaluation
set, whereas aggregation depends on the within-question ordering of rollouts. A model can
be well calibrated on average and still fail to rank its own parallel attempts. Consistent
with this, the curves separate on AIME24, AIME25, and OlympiadBench, where rollouts
disagree more often, and converge on MATH-500, where accuracy is near ceiling and the
oracle gap is small.

\end{document}